\documentclass[10pt,twocolumn,letterpaper]{article}

\usepackage{wacv}
\usepackage{times}
\usepackage{epsfig}
\usepackage{graphicx}
\usepackage{amsmath}
\usepackage{amssymb}
\usepackage{booktabs}
\usepackage{comment}
\usepackage{color}
\usepackage{xcolor}
\usepackage{booktabs}
\usepackage{multirow}
\usepackage{placeins}
\usepackage{rotating}

\usepackage[inline]{enumitem}

\usepackage[accsupp]{axessibility}

\DeclareMathOperator*{\argmax}{arg\,max}
\DeclareMathOperator*{\argmin}{arg\,min}

\wacvfinalcopy 

\usepackage[pagebackref=true,breaklinks=true,colorlinks,bookmarks=false]{hyperref}

\begin{document}

\title{DELS-MVS: Deep Epipolar Line Search for Multi-View Stereo}

\author{Christian Sormann \textsuperscript{1} \\
{\tt\small christian.sormann@icg.tugraz.at}
\and
Emanuele Santellani \textsuperscript{1}\\
{\tt\small emanuele.santellani@icg.tugraz.at}
\and 
Mattia Rossi \textsuperscript{2}  \\
{\tt\small mattia.rossi@sony.com}
\and
Andreas Kuhn \textsuperscript{2} \\
{\tt\small andreas.kuhn@sony.com}
\and Friedrich Fraundorfer \textsuperscript{1} \\
{\tt\small fraundorfer@icg.tugraz.at} 
\and 
\textsuperscript{1} Graz University of Technology \\
Institute of Computer Graphics and Vision
\and 
\textsuperscript{2} Sony Europe B.V. \\
R\&D Center - Stuttgart Laboratory 1
}

\maketitle
\thispagestyle{empty}

\begin{abstract}
\vspace{-1em}
We propose a novel approach for deep learning-based Multi-View Stereo (MVS). For each pixel in the reference image, our method leverages a deep architecture to search for the corresponding point in the source image directly along the corresponding epipolar line. We denote our method DELS-MVS: Deep Epipolar Line Search Multi-View Stereo. Previous works in deep MVS select a range of interest within the depth space, discretize it, and sample the epipolar line according to the resulting depth values: this can result in an uneven scanning of the epipolar line, hence of the image space. Instead, our method works directly on the epipolar line: this guarantees an even scanning of the image space and avoids both the need to select a depth range of interest, which is often not known a priori and can vary dramatically from scene to scene, and the need for a suitable discretization of the depth space. In fact, our search is iterative, which avoids the building of a cost volume, costly both to store and to process. Finally, our method performs a robust geometry-aware fusion of the estimated depth maps, leveraging a confidence predicted alongside each depth. We test DELS-MVS on the ETH3D, Tanks and Temples and DTU benchmarks and achieve competitive results with respect to state-of-the-art approaches.
\end{abstract}
\vspace{-1.5em}

\section{Introduction}

Multi-View Stereo (MVS) systems target the reconstruction of accurate and complete depth maps of multiple calibrated images capturing the same scene.
This is a long standing research topic in computer vision, as it is at the basis of any 3D reconstruction pipeline.
In recent years, deep learning based methods have been gaining a lot of attention in the field,
due to their more robust learned feature representations and matching similarity measures~\cite{mvsnet,patchmatchnet}.
\begin{figure}[t]
    \centering
    \includegraphics[width=0.9\columnwidth]{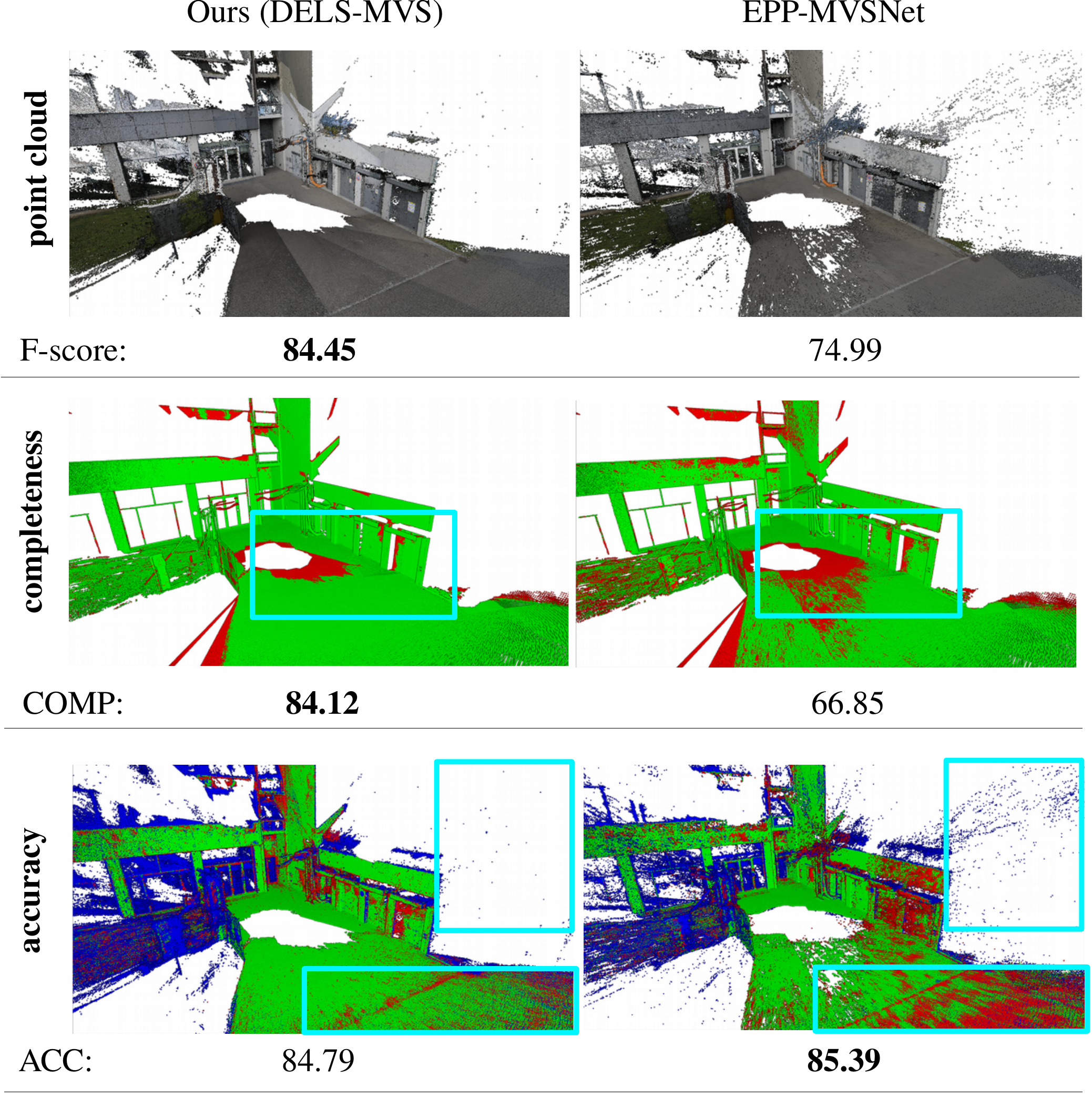}
    \caption{
    ETH3D high res~\cite{eth3d} benchmark visualization for the electro scene. Our point cloud is more complete and less noisy.
    }
    \label{fig_point_cloud_compare}
    \vspace{-1em}
\end{figure}
A large portion of these methods make use of depth-plane sweeping cost volumes, which have two distinct disadvantages.
The first is their large memory footprint, which typically results in high computational processing costs as well.
The second is the need to discretize the depth search space, which requires both to define a depth range of interest and to define
a discretization scheme for it.
Although the depth range can be inferred from the sparse reconstruction of a Structure from Motion (SfM) system,
the sparse nature of the reconstruction can lead to under or over estimate the depth range of the portion of the scene captured
by a specific image, thus preventing the reconstruction of some scene areas.
Moreover, some discretization strategies suit certain scenes better than others.
For objects close to the camera, a fine grained discretization toward the depth range minimum is preferable.
Conversely, objects farther in the distance can be reconstructed even with a coarser discretization.
The limitations behind depth discretization approaches becomes obvious when considering scenes containing multiple objects at
very different distances from the camera.
A better alternative is represented by a discretization of the inverse depth range~\cite{iter_mvs,inverse_depth_regression}, but this still calls for a depth range of interest.
In order to tackle both of these disadvantages, we propose a new method for MVS depth estimation denoted Deep Epipolar Line Search (DELS).
For each pixel in the reference image, we search for the matching point in the source image by exploring the corresponding epipolar line.
The search procedure is carried out iteratively and utilizes bi-directional partitions of the epipolar line around the current estimate to update the predicted matching point.
These updates are guided by a deep neural network, fed with features from both the reference and the source image.

Our approach has several advantages over those methods that first discretize a given depth range, chosen a priori, and later convert the resulting depth values into points or segments along the epipolar line, when looking for a match.
First, operating on the epipolar line allows our method to exploit the image information better.
In fact, due to the scene geometry and the relative pose between the reference and the source camera, a uniform discretization of the depth range could result in points clustered in a small segment of the epipolar line, thus preventing a correct match.

Second, our strategy avoids both the need to define a search depth range and the need for a depth discretization strategy customized for the scene content, as the epipolar line is explored dynamically.
Our method is iterative and adopts a coarse-to-fine approach that permits to scan the epipolar line efficiently.
This avoids the construction of a large fine grained depth cost volume.
Finally, our method estimates a depth at the reference image for each available source one and fuses them in a geometry-aware fashion,
using confidence measures estimated alongside the depth maps themselves.
These confidence measures can be leveraged also during point cloud construction, in order to filter outliers, thus leading to more accurate reconstructions.

In summary, our core contributions are the following:
\begin{enumerate*}[label=(\roman*)]
    \item a deep, iterative and coarse-to-fine depth estimation algorithm operating directly on the epipolar lines, thus avoiding the drawbacks of depth discretization, e.g. no specification of a depth range is needed
    \item a confidence prediction module and a geometry-aware fusion strategy that, coupled, permit a robust fusion of the multiple reference image depth maps from the different source images
    \item we verify the robustness of our method by evaluating on all the most popular MVS benchmarks, namely ETH3D \cite{eth3d}, Tanks and Temples \cite{tanksandtemples} and DTU \cite{dtu} and achieve competitive results.
\end{enumerate*}

\section{Related Work}
In order to estimate dense depth maps, MVS methods compute matching costs for a set of depth candidates at each pixel.
Initially, a common approach for candidate generation was the discretization of a depth range of interest, referred to as plane sweep.
The evaluation of the cost function at each candidate depth of each pixel
leads to a 3D cost volume that can be processed to extract the final depth map~\cite{plane_sweep_mvs}.
Later, the PatchMatch algorithm~\cite{patchmatch_algo} was adopted to generate candidate depth values~\cite{colmap_mvs,deepcmvs,gipuma,acmm}.
The main drawback of these MVS methods is their relying on hand engineered features and matching costs.
Instead, the recent deep MVS methods leverage more robust learned features.

Early deep MVS methods also used depth-plane sweeping cost volumes~\cite{mvsnet,deepmvs}, as these can be regularized by 3D convolutional neural networks. 
However, these methods are characterized by severe memory limitations.
Our proposed DELS-MVS leverages learned features, but it avoids constructing global cost volumes by employing a dynamic iterative approach guided by a neural network.
This differentiates DELS-MVS from those deep MVS methods relying on coarse-to-fine cost volumes~\cite{casmvs,cvp_mvsnet,ucsnet,unimvsnet,np_cvp_mvsnet} or cost volume aggregation schemes~\cite{epp_mvsnet,attmvsnet,transmvsnet,lanet} to limit the high memory and computational requirements associated to cost volume processing.
In particular, DELS-MVS working directly on the epipolar line, without any cost volume, sets it apart from EPP-MVSNet~\cite{epp_mvsnet}, a cost volume-based method that aggregates additional samples on the epipolar line to improve the quality of lower resolution cost volumes. 

Since DELS-MVS operates directly on the epipolar line in pixel space, it does not require a min./max. depth range, as the algorithm explores the epipolar line dynamically.
This differentiates DELS-MVS from EPP-MVSNet~\cite{epp_mvsnet} and previously proposed works leveraging iterative schemes~\cite{patchmatchnet,iter_mvs,ib_mvs,effi_mvs,raymvsnet} or cost volumes~\cite{casmvs,cvp_mvsnet,ucsnet,unimvsnet,np_cvp_mvsnet}, which generate hypothesis in (inverse) depth space and require a min./max. range as input.

Although an iterative depth estimation approach is adopted also in two-view stereo and SLAM methods based on the RAFT-concept~\cite{raft_stereo,droid_slam}, these methods require the construction of a full cost volume that prevents them from working at high resolutions. Instead, DELS-MVS can operate at full resolution, as it avoids full cost volumes. This also sets it apart from those MVS methods that update a pre-defined cost volume using a recurrent network~\cite{iter_mvs}.

Finally, operating on the epipolar line allows us to use additional information when fusing source views, which differentiates DELS-MVS from previous works~\cite{vismvsnet,patchmatchnet}. 
In particular, DELS-MVS proposes a geometry aware fusion strategy, which identifies and utilizes the source image for which the most accurate depth estimate is possible.
Moreover, our iterative scheme incorporates a mechanism to detect occluded regions via outer partitions.
\section{Algorithm}

The proposed system takes as input one reference image $\mathcal{R}$ and $N \geq 1$ source images
$\mathcal{S}^{0 \leq n \leq N - 1}$ along with the respective reference to source camera transformations
$T_{\mathcal{R} \rightarrow \mathcal{S}^n}$.
First, deep features are extracted from the reference and a given source image $\mathcal{S}^n$ by a convolutional neural network.
These features are then fed to our core algorithm, whose purpose is the estimation of the depth map $\mathcal{D}^n$
at the reference image.
For each reference image pixel, the goal of our algorithm is to estimate the residual between the actual pixel projections into the source image and our initial guess along the epipolar line.
This approach is introduced in
Section~\ref{sec_epi_depth_estimation}.
In order to avoid scale dependencies, our algorithm estimates the residuals via iterative classification steps, which are carried out in a coarse-to-fine fashion.
We name our algorithm Deep Epipolar Line Search (DELS), as the iterative classification resembles a search and leverages a deep neural network, referred to as Epipolar Residual Network (ER-Net). We describe the DELS algorithm, which represents the core of our DELS-MVS, as well as ER-Net, in Sections \ref{sec_dels_details} and \ref{sec_er_net_details}.

\begin{figure}[t]
    \centering
    \includegraphics[width=\columnwidth]{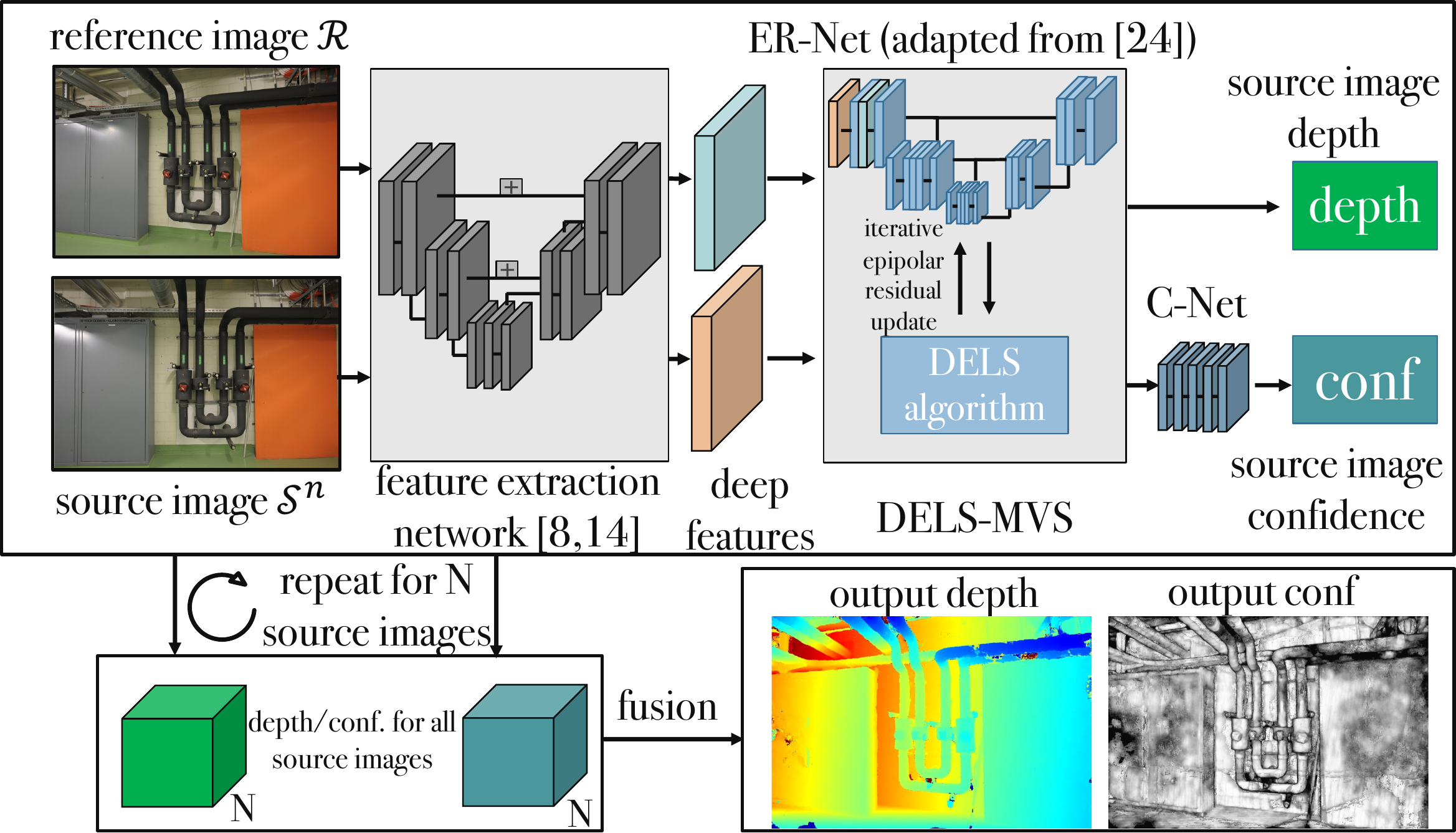}
    \caption{Overview visualization showing the individual steps performed in DELS-MVS. After extracting deep features, the deep epipolar line search yields output depth maps and confidence maps for each source image, later fused into a single output depth map.}
    \label{fig_arch_overview}
    \vspace{-1em}
\end{figure}

Finally, DELS-MVS also features a Confidence Network (C-Net) that associates a confidence map to the estimated depth
map $\mathcal{D}^n$.
This network is presented in Section~\ref{sec_conf_net_details} together with the procedure adopted to fuse all the $\mathcal{D}^{0 \leq n \leq N - 1}$
estimated depth maps at the reference image, utilizing each source, into a single depth map. 
We provide an overview visualization of DELS-MVS in Figure~\ref{fig_arch_overview}.
\subsection{Depth estimation via epipolar residual} \label{sec_epi_depth_estimation}

The DELS algorithm is initialiazed with a constant depth map $\mathcal{D}_0^n$, where the constant is computed as the average between
the minimum and maximum depth range obtained from a Structure from Motion (SfM) system.

\begin{figure}[t]
    \centering
    \includegraphics[width=0.8\columnwidth]{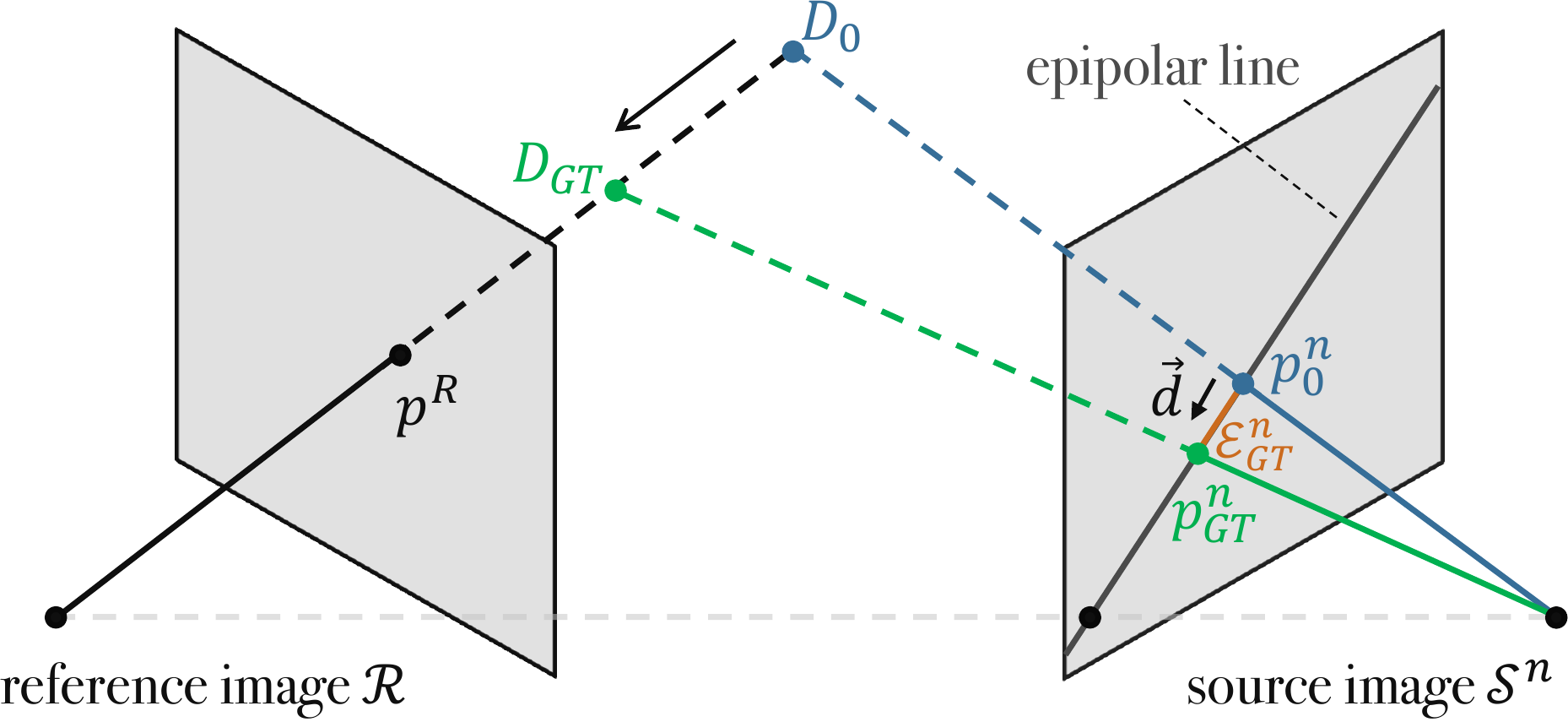}
    \caption{Visualization depicting depth estimation via epipolar residual prediction. Predicting the epipolar residual $\mathcal{E}^n_{{\text{GT}}}(x,y)$ permits to recover the ground truth depth.}
    \label{fig_epipolar}
    \vspace{-1.3em}
\end{figure}

Given a pixel $(x,y)$ in the reference image, let $p_0^n(x,y)$ be the projection of $(x,y)$ into the source image $\mathcal{S}^n$
according to the initial depth $\mathcal{D}_0^n(x,y)$.
Similarly, indicated with $\mathcal{D}_{\text{GT}}^n$ the sought Ground Truth depth map, let $p_{\text{GT}}^n(x,y)$ be the projection of $(x,y)$
into the source image $\mathcal{S}^n$ according to $\mathcal{D}_{\text{GT}}^n(x,y)$.
Both the projections lie on the same epipolar line, as depicted in Figure~\ref{fig_epipolar}.
Our target is the estimation of the one dimensional epipolar residual $\mathcal{E}^n_{{\text{GT}}} \in \mathbb{R}$ such that the following
relation holds true:
\begin{equation}
p^n_{\text{GT}}(x,y) = p^n_0(x,y) + \overrightarrow{d}(x,y) \mathcal{E}^n_{{\text{GT}}}(x,y)
\label{eq_point_from_res}
\end{equation}
where the unitary vector $\overrightarrow{d}(x,y) \in \mathbb{R}^2$ defines the epipolar line direction.
In fact, it is important to observe that an epipolar line does not have any direction a priori.
In particular, we orient $\overrightarrow{d}(x,y)$ such that moving consistently in its direction yields the depth hypothesis of $(x,y)$
to decrease monotonically.
For the sake of completness we observe that the following relation holds true:
\begin{equation}
    \left| \mathcal{E}^n_{{\text{GT}}}(x,y) \right| = \lVert p^n_{\text{GT}}(x,y) - p^n_0(x,y) \rVert_2.
\label{eq_epi_residual}
\end{equation}

Indicated with ${\mathcal{E}^n}(x,y)$ the epipolar residual estimated by DELS, Eq.~\eqref{eq_point_from_res} is used to correct the original estimate $p_0^n(x,y)$,
which yields $p^n(x,y)$.
This can now be converted into a depth value $\mathcal{D}^n(x,y)$ for the original pixel $(x,y)$ using the relative camera transformation
$T_{\mathcal{R} \rightarrow \mathcal{S}^n}$ from the reference to the source image, which is given.
This process is depicted in Figure~\ref{fig_epipolar}.

\subsection{Deep epipolar line search (DELS)} \label{sec_dels_details}

In the MVS scenario, the  baselines between the different source images and the reference one can vary a lot, both within
the same scene and not.
Moreover, depth maps can exhibit very different ranges depending on the specific scene: from very small ones, for the reconstruction of small objects,
to very large ones, for the reconstruction of outdoor scenes.
In most 3D reconstruction scenarios, the scene scale is not known a priori.
Overall, this makes the training of a network, regressing the epipolar error directly, a very challenging task.
For this purpose, we propose to re-cast the epipolar line residual estimation problem into an iterative and coarse-to-fine classification scheme.

\begin{figure}[t]
    \centering
    \includegraphics[width=\columnwidth]{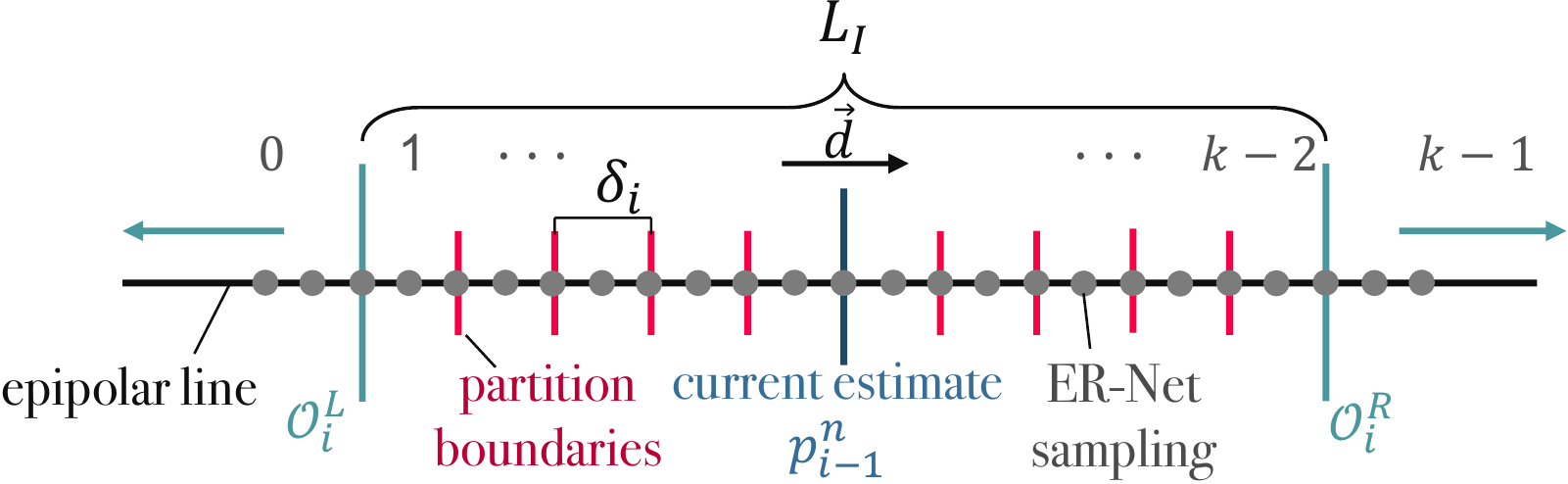}
    \caption{Visualization of the proposed partitioning of the epipolar line for classification, including sample positions of ER-Net.}
    \label{fig_sampling}
    \vspace{-1.3em}
\end{figure}

We denote the epipolar residual estimate after iteration $i$ as ${\mathcal{E}^n_i}(x,y)$ with i = $1,2,\ldots,I$.
In order to estimate the epipolar residual at a new iteration $i$, we segment the epipolar line into into $k$ partitions
$L = \{0,1...,k-1\}$ with $k$ even, as shown in Figure~\ref{fig_sampling}.
In particular we define two sets of partitions: the inner set $L_{\mathcal{I}} = \{1,...,k-2\}$ and the outer set
$L_{\mathcal{O}} = \{0,k-1\}$.
The inner set is centered at the previous iteration epipolar position $p_{i-1}^n(x,y)$ resulting from
${\mathcal{E}^n_{i-1}}(x,y)$ via Eq.~\eqref{eq_point_from_res}.
This set extends from a left border $\mathcal{O}^L_i(x,y) = -\frac{k-2}{2}\delta_i(x,y)$ to a right border $\mathcal{O}^R_i(x,y) = \frac{k-2}{2}\delta_i(x,y)$
where $\delta_i(x,y) \in \mathbb{R}^+$ denotes the size of each partition.
The partitions $0$ and $k-1$ of the outer set model the case where a match lies in a segment of the epipolar line not covered by the current inner partitions, i.e., outside the segment enclosed by $\mathcal{O}^L_i(x,y)$ and $\mathcal{O}^R_i(x,y)$.
Therefore, they provide us with an estimation on the direction to move towards in the next iteration.

\begin{figure*}[t]
    \centering
    \includegraphics[width=1.65\columnwidth]{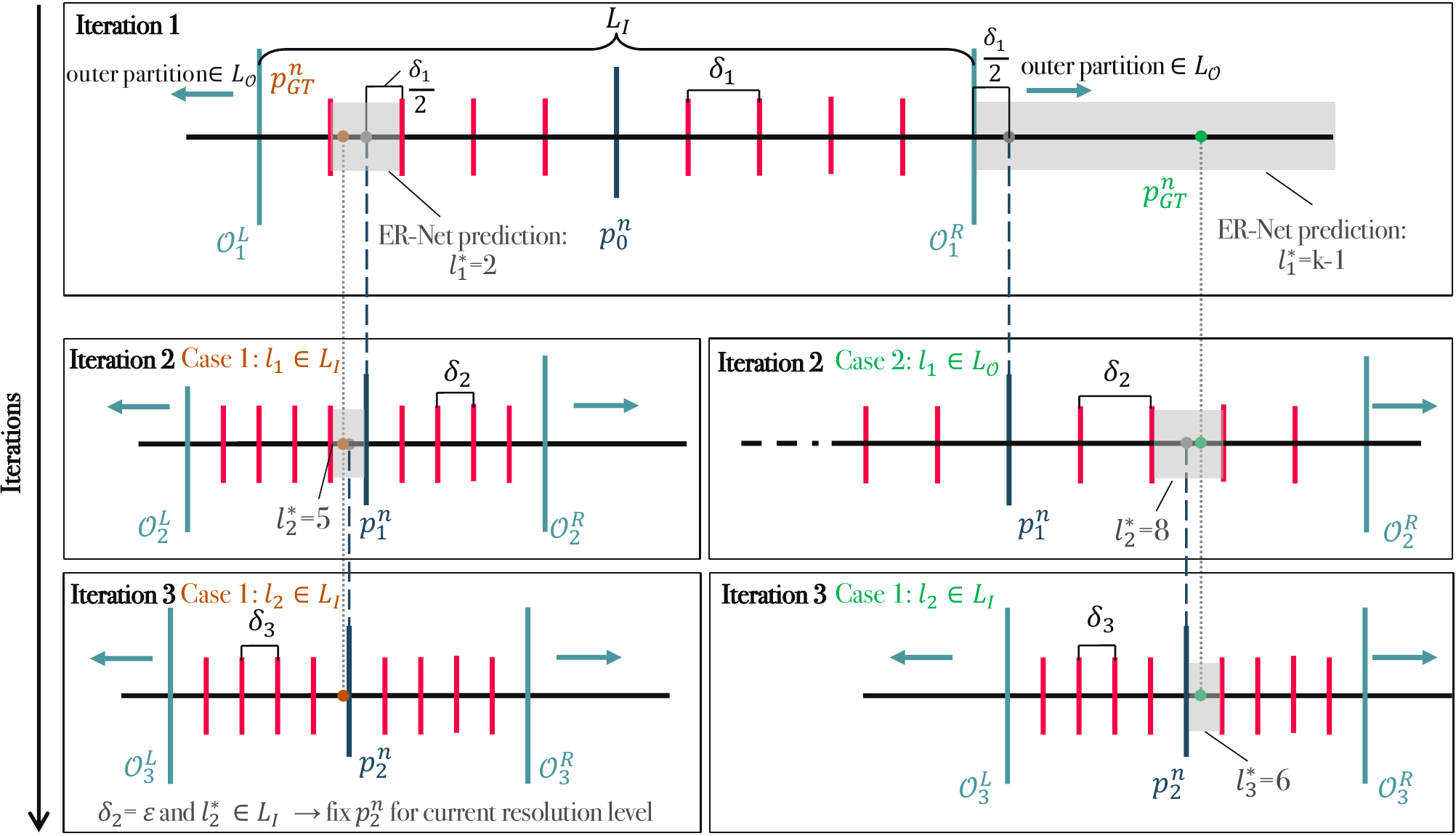}
    \caption{DELS algorithm visualization. In general, the algorithm differentiates between two cases: whether the predicted match on the epipolar line lies within the inner set $L_{\mathcal{I}}$ (case 1) or in the outer set $L_{\mathcal{O}}$ (case 2). This alters the partition width $\delta_i$ and position of the next estimate for the following iterations as described in Section~\ref{sec_dels_details}.}
    \label{fig_algorithm}
    \vspace{-1.2em}
\end{figure*}

We leverage a neural network to predict the partition $l_i^{\star}(x,y) \in L$ hosting the ground truth projection $p^n_{\text{GT}}(x,y)$.
Once the segment has been identified, a relative residual $e^n_{i}(x,y) \in \mathbb{R}$ can be computed and added to the previous iteration
residual ${\mathcal{E}^n_{i-1}}(x,y)$ to obtain the current iteration:
\begin{equation}
    {\mathcal{E}^n_{i}}(x,y) = {\mathcal{E}^n_{i-1}}(x,y) + e^n_{i}(x,y),
\end{equation}
with $ {\mathcal{E}^n_{0}}(x,y) = 0$.
In particular, we consider two scenarios:
\begin{itemize}
    \item Case 1: $l_i^{\star}(x,y) \in L_{\mathcal{I}}$ is selected, which predicts $p_{\text{GT}}^n(x,y)$ as located
    somewhere inside the finite partition $l_i^{\star}(x,y)$, hence $p_i^n(x,y)$ is placed at its mid point.
    \item Case 2: the partition $l_i^{\star}(x,y) \in L_{\mathcal{O}}$ is selected, which provides only a general hint regarding the search
    direction for $p_{\text{GT}}^n(x,y)$, hence $p_i^n(x,y)$ is placed at $\mathcal{O}^L_i(x,y) - \delta_i(x,y)/2$ and $\mathcal{O}^R_i(x,y) + \delta_i(x,y)/2$
    for $l_i^{\star}(x,y) = 0$ and $k-1$, respectively. In particular, this strategy permits to compensate for incorrect predictions at the
    previous iterations, as some overlap is maintained between the inner partitions of the current and next iteration.
\end{itemize}
\vspace{-0.3em}
The two scenarios are summarized in Figure~\ref{fig_algorithm}, where three DELS iterations are sketched.
Overall, for a predicted partition $l_i^{\star}(x,y) \in L$, the outlined update strategy is equivalent to define the relative residual as follows:
\begin{equation}
    e^n_i(x,y) = \mathcal{O}^L_i(x,y) + (l_i^{\star}(x,y) - 1)\delta_i(x,y) + \frac{\delta_i(x,y)}{2}.
    \label{eq_epi_shift}
\end{equation}

DELS coarse-to-fine scheme updates the partition width $\delta_i$ at each iteration, differentiating again between the Case 1 and 2.
In fact, in the first scenario, the partition width is halved at the beginning of the iteration $i+1$, while in the second scenario it is left
untouched:
\begin{equation}
    \delta_{i+1}(x,y) =
    \begin{cases}
        \max(\frac{1}{2} \delta_i(x,y), \epsilon) & l_i^{\star}(x,y) \in L_{\mathcal{I}} \\
        \delta_i(x,y) & l_i^{\star}(x,y) \in L_{\mathcal{O}}
    \end{cases},
\label{eq_part_update}
\end{equation} 
where $\epsilon \in \mathbb{R}^+$ is a lower bound on the partition width.
Finally, DELS adopts a multi-resolution strategy with three levels $j=0,1,2$ representing full, half and quarter-resolution, respectively.
The number of iterations at level $j$ is denoted $I_j$.
The level $j$ is initialized with the up-scaled epipolar residual from the previous scale $j+1$, that is $\mathcal{E}^n_{j} = NN_{\uparrow}(\mathcal{E}^n_{j+1}) \cdot 2$ where $NN_{\uparrow}$ is the nearest neighbour upsampling operator.
Further, when the partition width lower bound for a given level $\epsilon_j$ is reached and the selected partition is contained in the inner set, the iterations are interrupted for a given pixel, as the maximum achievable precision for the epipolar residual is considered achieved.
This behaviour is exemplified in Figure~\ref{fig_algorithm} (iteration $3$, left side).
\subsection{Epipolar Residual Network} \label{sec_er_net_details}

The classification task at each DELS iteration is carried out by a deep neural network denoted Epipolar Residual Network (ER-Net).
The network is fed with the feature maps $F^{\mathcal{R}}$ and $F^{\mathcal{S}^n}$ extracted from the
reference and source images $\mathcal{R}$ and $\mathcal{S}^n$, respectively, using the architecture proposed in~\cite{casmvs,feature_pyramid_net}. 
Additionally, ER-Net receives the previous iteration epipolar residual map $\mathcal{E}^n_{i-1}$: this permits, for each reference image
pixel $(x,y)$, to sample the features $F^{\mathcal{S}^n}$ located on the epipolar line around the latest estimate $p_{i-1}^n(x,y)$.
To this purpose, we incorporate deformable convolutions~\cite{deform_convsv2} inside a U-Net-like architecture adapted from~\cite{ib_mvs}.
In particular, at the top level of the U-Net, the deformable kernel is spread along the epipolar line and aligned such that it overlaps
with samples equally spaced at distance $\delta_i(x,y) / 2$ from each other, before and after $p_{i-1}^n(x,y)$.
In this setup, the samples in the inner set fall at the beginning, mid and end points of each partition.
The sampling scheme is represented by the gray dots in Figure~\ref{fig_sampling}.
The kernel size is chosen to cover all the samples in the inner set and extend further to the outer set.
In particular we set $k=12$ and the kernel size to five, yielding $5 \times 5 = 25$ samples.
For the U-Net lower resolution levels, we keep the same partition widths and kernel size as at the highest level, as this increases the
network receptive field.

The multi-channel output layer of the network output is fed to a softmax activation layer, which yields a probability for each one of
the $k$ partitions $l_i(x,y) \in L$:
\begin{equation}
    P(l_i,x,y) = \text{ER-Net}(F^{\mathcal{R}}(x,y), F^{\mathcal{S}^n}(x,y), {\mathcal{E}^n_{i-1}}(x,y)).
\end{equation}
During inference, the partition with the maximum softmax probability is chosen:
\begin{equation}
l_i^{\star}(x,y) = \argmax_{l_i} P(l_i,x,y).
\label{eq_l_star}
\end{equation}
As anticipated in Sec.~\ref{sec_dels_details}, we adopt ER-Net within a multi-resolution scheme involving full, half and quarter resolution, but we do not share parameters between the levels.

\subsection{Confidence Network} \label{sec_conf_net_details}

Our method computes $N$ depth maps at the reference image, each one computed using a different source image.
This poses the problem of how to fuse them into a single depth map taking advantage of all the estimated ones,
as some reference image areas may be visible in one source image while not in another.
For this purpose, we introduce a Confidence Network (C-Net), which is used to assign a confidence map $C^n$ to each estimated depth map $D^n$: the confidence maps are then used to guide the fusion of the multiple available depth maps.

At each level $j$ of our multi-resolution scheme, we compute a map resembling the pixel-wise entropy of the partition probabilities,
but taking into account its evolution over the DELS iterations:
\begin{equation}
 \Tilde{C}^n_{j}(x,y) = \frac{1}{I_{j}} \sum_{i=1}^{I_{j}} \sum_{l_i \in L} -P(l_i,x,y) \log(P(l_i,x,y)).
\end{equation}
We fed C-Net with $\Tilde{C}^n = NN_{\uparrow}(\Tilde{C}^n_{2}) + NN_{\uparrow}(\Tilde{C}^n_{1}) + \Tilde{C}^n_{0}$,
which takes into account the different resolution levels.
Moreover, we set $\Tilde{C}(x,y)$ to the maximum entropy value in locations where the estimated partition in the last iteration has the property $l_I^{\star}(x,y) \in L_{\mathcal{O}}$.
This incorporates the knowledge that predictions in the last iteration associated to the outer set $L_{\mathcal{O}}$ yield unreliable depth estimates, indicating that an occlusion is present.
C-Net architecture consists of $4$ convolution layers (kernel size $3$, leaky ReLU activations) and a sigmoid activated output yielding $C^n$ in range $[0,1]$.
\vspace{-0.1em}
\subsection{Geometry-aware multi-view fusion} \label{sec_geom_aware_fusion}

At each pixel $(x,y)$, a set of $N$ depth candidates is available, one for each source image.
The depth candidate $D^n(x,y)$ is associated to the center of the final epipolar line partition predicted by the DELS algorithm,
with the partition length corresponding to a depth range of size $\Delta D^n(x,y) \in \mathbb{R}$ in the scene.
It is crucial to observe that, despite the epipolar line partitions sharing the same length across the different source images,
the corresponding depth ranges $\Delta D^n(x,y)$ vary from source image to source image, depending on the geometry relating the source cameras to the reference one.
In our method, we leverage this key fact and propose a geometry-aware fusion scheme utilizing learned confidences. 
In particular, depth candidates whose associated partition has both a sufficient confidence and a narrow depth range are preferred, as they correspond to reliable and fine grained depth predictions.
Thus, our fusion process proceeds as follows.
First, we determine the set $N_{\omega}$ of candidates with a confidence larger than threshold $\omega$. 
Then, we select the candidate with the smallest depth partition size, that is ${\mathcal{D}^{n^\star}(x,y)}$ where
$n^{\star} = \argmin_{n} \Delta D^n(x,y)$ with $n \in N_{\omega}$ (for $N_{\omega}$ empty, selecting the one with highest confidence).
The final depth is the average of all candidates within a relative distance threshold $\eta$ from ${\mathcal{D}^{n^{\star}}(x,y)}$.
For this purpose, we introduce the mask:
\begin{equation}
 M^n(x,y) = 
 \begin{cases}
        1 & \frac{\lvert \mathcal{D}^{n^{\star}}(x,y) - \mathcal{D}^{n}(x,y) \rvert}{\mathcal{D}^{n^{\star}}(x,y)} < \eta \\
        0 & \text{else} .
\end{cases}
\label{eq_conf_net_threshold}
\end{equation}
The final output depth value can be expressed as follows:
\begin{equation}
\mathcal{D}(x,y) = \frac{1}{\sum_{n=0}^{N-1} M^n(x,y)} \sum_{n=0}^{N-1} M^n(x,y) {\mathcal{D}^n(x,y)}.
\label{eq_depth_fusion}
\end{equation}
\section{Experimental evaluation and Training}

In the following sections we introduce our training procedure and present DELS-MVS evaluation results on multiple benchmarks.
Furthermore, we motivate contributions and parameters of DELS-MVS with experiments.

\subsection{Training procedure}

The proposed method is implemented using PyTorch~\cite{pytorch} and trained utilizing the ADAM optimizer~\cite{adam_optimizer} with batch size $1$.
\subsubsection{ER-Net}

For our experiments on DTU~\cite{dtu}, the network is trained on the pre-processed DTU training set of~\cite{mvsnet} with learning rate $10^{-4}$.
When evaluating on Tanks and Temples~\cite{tanksandtemples} and ETH3D~\cite{eth3d}, we additionally fine tune the network on the Blended MVS dataset~\cite{blended_mvs} with learning rate $10^{-5}$.

At each training iteration, we pick a random source image and initialize the DELS algorithm with a constant
random depth map, similarly to \cite{patchmatchnet} and \cite{bi3d_stereo,ib_mvs}, respectively.
A training step involves $I_j$ iterations of the DELS algorithm, depending on the resolution level $j=0,1,2$.
The ground truth line partition at iteration $i$ of DELS is computed from the ground truth relative epipolar residual
$e^n_{i_{\text{GT}}}(x,y) = \mathcal{E}^n_{{\text{GT}}}(x,y) - {\mathcal{E}^n_{i-1}}(x,y)$ as follows:
\begin{equation}
    {(l_i)}_{\text{GT}}(x,y) = \sum_{g=0}^{k-2} [ e^n_{i_{\text{GT}}}(x,y) > (\mathcal{O}^L_i(x,y) + g \delta_i(x,y)) ],
\end{equation}
with $[...]$ being the Iverson bracket, which yields $1$ if the condition is true and $0$ otherwise.
We apply a Cross Entropy (CE) loss between the softmax probabilities $P(l_i,x,y)_{l_i \in \mathcal{L}}$ and the one hot encoding
of the ground truth partition ${(l_i)}_{\text{GT}}(x,y)$.
The total loss is the sum of the cross entropy losses over the multiple iterations:
\begin{equation}
    \text{Loss}^{\text{ER}}_j(x,y) = \sum_{i=1}^{I_j} \text{CE}(P(l_i,x,y), {(l_i)}_{\text{GT}}(x,y)).
\end{equation}
During training no gradient flows to the previous iteration.
Moreover, the training includes the different resolution levels progressively.
Specifically, we start by training the coarsest level $j=2$ alone for $4$ epochs, then we train $j=2$ and $j=1$ for $3$ epochs jointly,
finally we train all the three levels $j=2, j=1, j=0$ together for $10$ additional epochs.
When fine tuning on the Blended MVS~\cite{blended_mvs} dataset, this last step is extended with $5$ additional epochs.
\subsubsection{C-Net}

In order to train our confidence network C-Net, we define a binary ground truth confidence map $C^n_{\text{GT}}$ where $1$ and $0$ indicate
valid and non valid depth measurements, respectively.
We define the ground truth confidence at $\mathcal{S}^n$ as:
\begin{equation}
    C^n_{\text{GT}}(x,y) = 
    \begin{cases}
        1 & \lVert p^{n}_{\text{GT}}(x,y) - p^{n}(x,y) \rVert_2 < \gamma \\
        0 & else
    \end{cases}
\end{equation}
where $p^{n}_{\text{GT}}(x,y)$ and $p^{n}(x,y)$ are the epipolar line projections of the reference image pixel $(x,y)$ according to
its ground truth depth and the depth estimated by DELS, respectively.
The confidence network is then trained with Binary Cross Entropy (BCE) loss between the predicted confidence $C^n$ and its ground truth defined above:
\begin{equation}
    \text{Loss}^{\text{C}} = BCE(C^n(x,y), C^n_{\text{GT}}(x,y)).
\end{equation}
C-Net is trained separately, after ER-Net training has been completed.
It is trained for $2$ epochs on the DTU~\cite{dtu} dataset, adopting a learning rate equal to $10^{-5}$ and $\gamma = 0.6$.
C-Net training also adopts source images picked at random.
\begin{table*}[t]
\centering
\small
\begin{tabular}{@{}llcccccc@{}}
 \toprule
 & & \multicolumn{3}{c}{\textbf{Training Set}} & \multicolumn{3}{c}{\textbf{Testing Set}}\\
 \cmidrule{3 - 8} 
 & \textbf{method} & F [\%] & AC [\%] & CO [\%] & F [\%] & AC [\%] & CO [\%] \\ \midrule
 \multirow{5}{*}{\begin{sideways}\textbf{classical}\end{sideways}} 
 & ACMM~\cite{acmm} & 78.86 & 90.67 & 70.42 & 80.78 & 90.65 & 74.34  \\
 & COLMAP~\cite{colmap_mvs} & 67.66 & \underline{91.85} & 55.13 & 73.01 &  \underline{91.97} & 62.98  \\
 & PCF-MVS~\cite{pcf_mvs} & 79.42 & 84.11 & 75.73 & 80.38 & 82.15 & 79.29 \\
 & MAR-MVS~\cite{mar_mvs} & 79.21 & 81.98 & 77.19 & 81.84 & 80.24 & 84.18 \\
 & DeepC-MVS~\cite{deepcmvs} & \underline{84.81} & 90.37 &  \underline{80.30} &  \underline{87.08} & 89.15 &  \underline{85.52} \\ \hline
 \multirow{8}{*}{\begin{sideways}\textbf{learned}\end{sideways}} 
 & IB-MVS~\cite{ib_mvs} & 71.21 & 75.21 & 69.02 & 75.85 & 71.64 & 82.18  \\
 & Iter-MVS~\cite{iter_mvs} & 71.69 & 79.79	 & 66.08 & 80.06 & 84.73 & 76.49 \\
 & PVSNet~\cite{pvsnet} & 67.48 & / & / & 72.08 & 66.41 & 80.05 \\
 & EPP-MVSNet~\cite{epp_mvsnet} & 74.00 & 82.76 & 67.58 & 83.40 & 85.47 & 81.79 \\
 & PatchMatch-Net~\cite{patchmatchnet} & 64.21 & 64.81 & 65.43 & 73.12 & 69.71 & 77.46 \\
 & Vis-MVSNet~\cite{vismvsnet} & 72.77 & \textbf{83.32} & 65.53 & 83.46 & \textbf{86.86} & 80.92	 \\
 & Ours (DELS-MVS) & \textbf{79.15} & 79.71 & \textbf{79.49} & \textbf{85.41} & 81.44 & \textbf{90.05} \\
 \bottomrule
\end{tabular}
\caption{ETH3D High-Res~\cite{eth3d} results, showing F-score, accuracy and completeness (percentage measurement using 2cm tolerance, higher is better). Best classical category results are underlined and best learned bold.}
\label{tbl_eth_high_res}
\end{table*}
\begin{table*}[t]
\centering
\small
\begin{tabular}{@{}llcccccc@{}}
 \toprule
 & & \multicolumn{3}{c}{\textbf{Intermediate Set}} & \multicolumn{3}{c}{\textbf{Advanced Set}}\\
 \cmidrule{3 - 8} 
 & \textbf{method} & F [\%] & PR [\%] & RC [\%] & F [\%] & PR [\%] & RC [\%] \\ \midrule
 \multirow{4}{*}{\begin{sideways}\textbf{classical}\end{sideways}} 
 & ACMM~\cite{acmm} & 57.27 & 49.19 & \underline{70.85} & 34.02 & 35.63 & 34.90   \\
 & COLMAP~\cite{colmap_mvs} & 42.14 & 43.16 & 44.48 & 27.24 & 33.65 & 23.96    \\
 & PCF-MVS~\cite{pcf_mvs} & 55.88 & 49.82 & 65.68 & \underline{35.69} & 37.52 & \underline{35.36}  \\
 & DeepC-MVS~\cite{deepcmvs} & \underline{59.79} & \underline{59.11} & 61.21 & 34.54 & \underline{40.68} & 31.30  \\ \hline
 \multirow{14}{*}{\begin{sideways}\textbf{learned}\end{sideways}} 
 & Att-MVS~\cite{attmvsnet} & 60.05 & \textbf{61.89} & 58.93 & 31.93 & \textbf{40.58} & 27.26 \\
 & CasMVSnet~\cite{casmvs} & 56.84 & 47.62 & 74.01 & 31.12 & 29.68 & 35.24 \\
 & VisMVSnet~\cite{vismvsnet} & 60.03 & 54.44 & 70.48 & / & / & / \\
 & EPP-MVSNet~\cite{epp_mvsnet} & 61.68 & 53.09 & 75.58 & 35.72 & 40.09 & 34.63 \\
 & IB-MVS~\cite{ib_mvs} & 56.02 & 47.71 & 72.64 & 31.96 & 27.85 & 41.48   \\
 & IterMVS~\cite{iter_mvs} & 56.94 & 47.53 & 74.69 & 34.17 & 28.70 & 44.19 \\
 & PatchMatch-Net~\cite{patchmatchnet} & 53.15 & 43.64 & 69.37 & 32.31 & 27.27 & 41.66 \\
 & GBi-Net~\cite{gbi_net} & 61.42 & 54.48 & 71.25 & 37.32 & 30.58 & \textbf{48.83} \\
 & Effi-MVS~\cite{effi_mvs} & 56.88 & 47.53 & 71.58 & 34.39 & 32.23 & 41.90 \\
 & NP-CVP-MVSNet\cite{np_cvp_mvsnet} & 59.64 & 47.05 & \textbf{84.69} & / & / & / \\
 & TransMVSNet~\cite{transmvsnet} & 63.52 & 55.14 & 76.73 & 37.00 & 33.84 & 44.29 \\
 & Uni-MVSNet~\cite{unimvsnet} & \textbf{64.36} & 57.54 & 73.82 & \textbf{38.96} & 33.76 & 47.22 \\
 & RayMVSNet~\cite{raymvsnet} & 59.48 & 53.21 & 69.21 & / & / & / \\
 & Ours (DELS-MVS) & 63.08 & 55.56 & 74.03 & 37.81 & 34.49 & 45.71  \\
 \bottomrule
\end{tabular}
\caption{Tanks and Tempels~\cite{tanksandtemples} results, showing F-score, precision and recall (percentage measurement, higher is better). Best classical category results underlined and best learned category results are bold.}
\label{tbl_tanks}
\end{table*}
\begin{table}[t]
\centering
\small
\begin{tabular}{@{}lccc@{}}
 \toprule
 & \multicolumn{3}{c}{\textbf{metrics [mm]}} \\
 \cmidrule{2 - 4} 
 \textbf{method} & AVG & AC & CO \\ \midrule
 CasMVSnet~\cite{casmvs} & 0.355 & 0.325 & 0.385  \\
 BP-MVSNet~\cite{bp_mvsnet} & 0.327 & 0.333 & 0.320 \\
 VisMVSnet~\cite{vismvsnet} & 0.365 & 0.369 & 0.361  \\
 EPP-MVSNet~\cite{epp_mvsnet} & 0.355 & 0.413 & 0.296  \\
 IB-MVS~\cite{ib_mvs} & 0.321 & 0.334 & 0.309  \\
 IterMVS~\cite{iter_mvs} & 0.363 & 0.373 & 0.354   \\
 PatchMatch-Net~\cite{patchmatchnet} & 0.352 & 0.427 & 0.277 \\
 GBi-Net~\cite{gbi_net} & \textbf{0.289} & \textbf{0.315} & \textbf{0.262} \\
 Effi-MVS~\cite{effi_mvs} & 0.317 & 0.321 & 0.313 \\
 NP-CVP-MVSNet\cite{np_cvp_mvsnet} & 0.315 & 0.356 & 0.275 \\
 TransMVSNet~\cite{transmvsnet} & 0.305 & 0.321 & 0.289 \\
 RayMVSNet~\cite{raymvsnet} & 0.330 & 0.341 & 0.319 \\
 Uni-MVSNet~\cite{unimvsnet} & 0.315 & 0.352 & 0.278 \\
 Ours (DELS-MVS) & 0.313 & 0.342 & 0.284   \\
 \bottomrule
\end{tabular}
\caption{DTU~\cite{dtu} results, showing completeness, accuracy, average of both (distance measurement, lower is better).}
\label{tbl_dtu}
\vspace{-1em}
\end{table}
\begin{table}[t]
\small
\centering
\setlength{\tabcolsep}{4pt}
\begin{tabular}{@{}ccccccccc@{}}
 \toprule
 & & & & & & \multicolumn{3}{c}{\textbf{metrics [\%]}} \\
 \cmidrule{7 - 9} 
 & PU & fusion & $\beta$ & $I$ & $N$ & F & AC & CO \\ \midrule
 1 & $\times$ & avg. & $\times$ & 4,2,2 & 2 & 52.80 & 53.85 & 53.73   \\
 2 & $\checkmark$ & avg. & $\times$ & 4,2,2 & 2 & 61.88 & 67.45 & 59.12   \\
 3 & $\checkmark$ & outer part. mask & $\times$ & 4,2,2 & 2 & 73.99 & 74.47 & 74.09 \\
 4 & $\checkmark$ & outer part. mask & $\times$ & 4,2,2 & 6 & 77.09 & 76.12 & 78.59 \\
 5 & $\checkmark$ & non geom. aware & $\times$ & 4,2,2 & 6 & 78.18 & 76.06 & 81.00 \\
 6 & $\checkmark$ & geom. aware & $\times$ & 4,2,2 & 6 & 78.76 & 77.03 & \textbf{81.16} \\
 7 & $\checkmark$ & geom. aware & $\times$ & 3,1,1 & 6 & 74.14 & 71.52 & 77.64 
 \\ 
 8 & $\checkmark$ & geom. aware & $\checkmark$ & 4,2,2 & 6 & \textbf{79.15} & \textbf{79.71} & 79.49   \\
 \bottomrule
\end{tabular}
\caption{Architectural experiments on the ETH3D training set~\cite{eth3d}, where PU denotes whether the dynamic partiton update is applied. The second column describes the technique for fusing the $N$ source depth maps. $\beta$ denotes whether the output confidence was used to filter point cloud results.  $I_{2}, I_{1}, I_{0}$ are the iterations.
}
\label{tbl_ablation}
\vspace{-1em}
\end{table}
\subsection{Benchmark datasets}
We evaluate the proposed DELS-MVS on the DTU \cite{dtu}, ETH3D~\cite{eth3d} and Tanks and Temples~\cite{tanksandtemples} benchmarks. 
These evaluate the accuracy and completeness of the point clouds reconstructed by an MVS algorithm.
For DTU~\cite{dtu}, these two metrics are expressed as average distances in $mm$.
Instead, ETH3D~\cite{eth3d} and Tanks and Temples~\cite{tanksandtemples} express these two metrics as percentages,
with Tanks and Temples~\cite{tanksandtemples} using the terms precision and recall for accuracy and completeness, respectively.
In order to generate the DELS-MVS point clouds, we adopt the dynamic fusion approach of~\cite{dhcrmvsnet}.
Moreover, we leverage our learned confidence to filter out outliers during the fusion.
Specifically, in each depth map $\mathcal{D}$, we filter out those pixels with a confidence $C(x,y)$ below a threshold $\beta$,
where $C$ is computed simply by replacing $D^n$ with $C^n$ in Eq.\eqref{eq_depth_fusion}.
We set $\beta=0.05$ for DTU, $\beta=0.1$ for Tanks and Temples and $\beta=0.2$ for ETH3D.
The DELS-related parameters $\delta_1$ and $\epsilon_{0}, \epsilon_{1}, \epsilon_{2}$ of Sec.~\ref{sec_dels_details} are set to $4$ and $0.5, 1.0, 2$, respectively.
The confidence-related parameters $\eta$ and $\omega$ in Sec.~\ref{sec_conf_net_details} are set to $0.01$ and $0.2$.
In all our tests we consider $N=6$ source images.
Concerning runtime and memory usage, on DTU~\cite{dtu} a $1600\times1184$ depth map requires 2.77s and 6.0GB of memory.
On ETH3D~\cite{eth3d}, a $1984\times1312$ depth map requires 3.63s and 7.5GB.
Finally, on Tanks and Temples~\cite{tanksandtemples}, a $1920\times1056$ depth map requires 2.87s and 6.2GB.
\subsection{ETH3D results}

We present results on the high resolution training and test sets in Table~\ref{tbl_eth_high_res}.
The proposed DELS-MVS shows state-of-the-art results, yielding the best results among the learning-based methods on both the training and test sets.
It should be noted that DELS-MVS was never trained on the ETH3D training set. 
Concerning the test set, DELS-MVS yields similar results to the current state-of-the-art classical method DeepC-MVS~\cite{deepcmvs}.
In Figure~\ref{fig_point_cloud_compare}, we show point cloud visual comparisons using benchmark visualizations. 
It can be observed that our method exhibits superior completeness and less noise compared to EPP-MVSNet~\cite{epp_mvsnet}.
\subsection{Tanks and Temples results}
\begin{figure}[t]
    \centering
    \includegraphics[width=0.9\columnwidth]{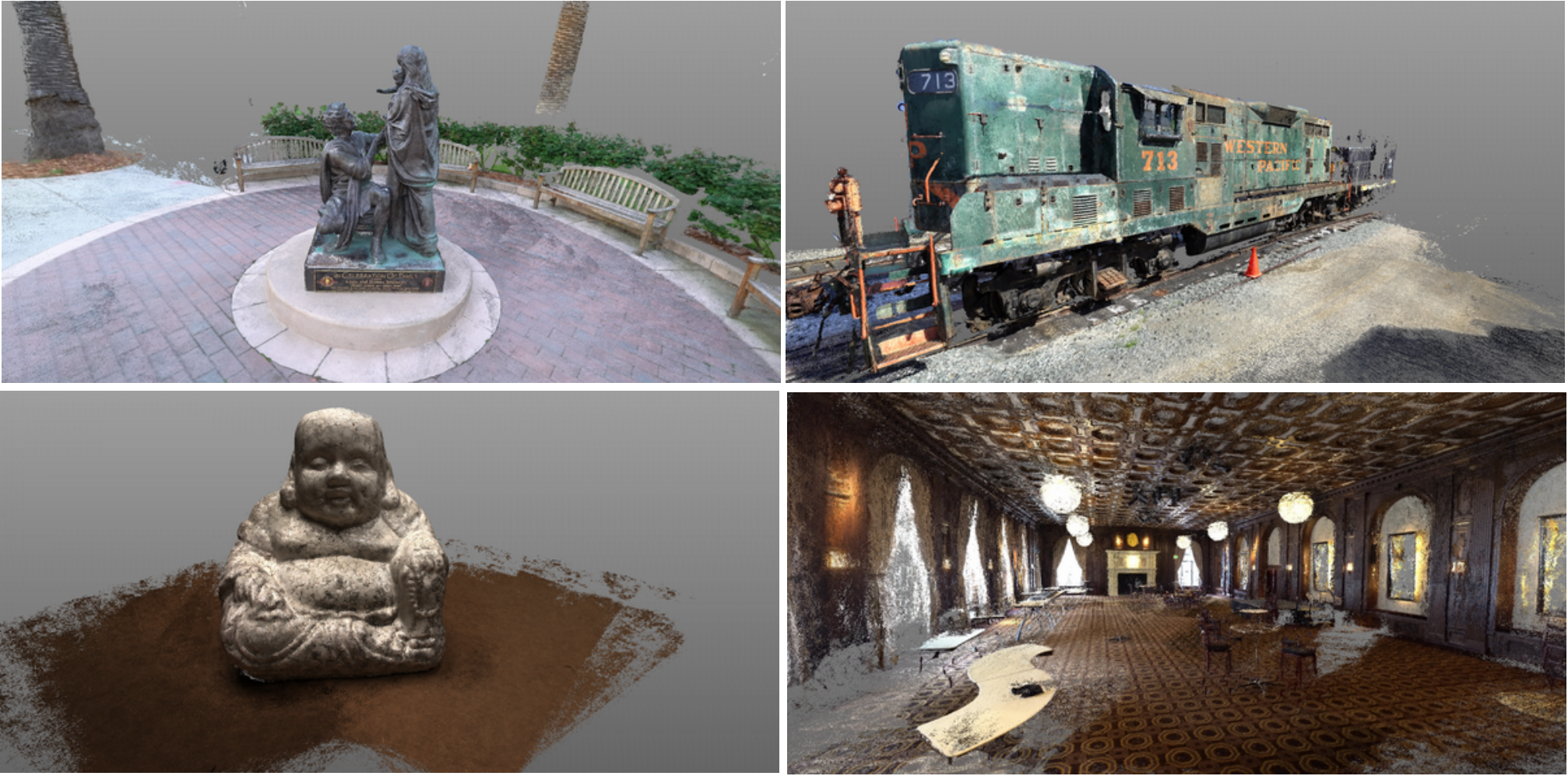}
    \caption{Point cloud results of DELS-MVS on DTU~\cite{dtu} and Tanks and Temples~\cite{tanksandtemples}.}
    \label{fig_point_cloud_results_dtu_tt}
    \vspace{-1.3em}
\end{figure}

We report the benchmark results from the intermediate and advanced sets in Table~\ref{tbl_tanks}.
DELS-MVS results are competitive on both subsets, especially on the advanced set, where it achieves results close to Uni-MVSNet~\cite{unimvsnet}. 
The advanced set contains less constrained camera poses and, compared to intermediate, it consists of more challenging reconstruction scenarios,
as suggested by the lower avg. scores of all methods.
We provide visual results in Fig.~\ref{fig_point_cloud_results_dtu_tt}.
\subsection{DTU results}

The evaluation results for DTU~\cite{dtu} are reported in Table~\ref{tbl_dtu}.
In terms of AVG score, which summarizes the overall reconstruction quality, DELS-MVS exhibits improved results compared to recent methods, such as IterMVS~\cite{iter_mvs} and EPP-MVSNet~\cite{epp_mvsnet}.
Moreover, DELS-MVS achieves AVG results which are competitive with other state-of-the-art methods, Uni-MVSNet~\cite{unimvsnet}
and Trans-MVSNet~\cite{effi_mvs}.
\subsection{Architectural experiments}

In this section we discuss ablation experiments carried out on the ETH3D~\cite{eth3d} high res training set and report them in Table~\ref{tbl_ablation}.
First, we evaluate the influence of our partition update strategy in Eq.~\eqref{eq_part_update}. Specifically, row 1 reports the results of a naive version of our algorithm where the partition size is updated exclusively when proceeding to the next level of the multi-resolution scheme.
It can be observed that, in row 2, the proposed dynamic update (Sec.~\ref{sec_dels_details}) leads to a significant performance improvement.

Next, we evaluate the effectiveness of the employed outer partitions in modeling unreliable or occluded estimates (Sec.~\ref{sec_geom_aware_fusion}).
In particular, in row 2 we report the results of a fusion variant that, at each pixel, averages the depth candidates obtained from the different source images regardless of the associated partition.
In row 3, instead, the fusion average masks out those depth candidates associated to an outer partition.
By comparing rows 2 and 3 it can be observed that the masking based on the outer partitions yields a performance increase, thus confirming the validity of the outer partition concept.
We evaluate also how the number of available source views $N$ affects DELS-MVS, and observe that the results improve when passing from $N=2$ to $N=6$ in rows 3 and 4, respectively.

Regarding our proposed confidence measure, row 5 shows that selecting the depth candidates with the highest confidence ${\mathcal{D}^{n^\star}(x,y)}$ in the fusion stage yields a further gain.
Row 6 confirms that even better results can be obtained when employing our proposed geometry aware fusion, which is made possible by DELS-MVS operating directly on the epipolar line.
Row 7 shows that a lower number of iterations leads to worse results, as expected.
Finally, in the last row we observe that an additional filtering of the final point clouds
with our learned confidence measure leads to the best results.
\section{Conclusion}

We proposed DELS-MVS, a novel MVS method that leverages a deep neural network to carry out a matching search directly on the source image epipolar line.
After estimating a dense depth map at the reference image for each available source, DELS-MVS employs a geometry-aware strategy to fuse them into a single depth map 
utilizing a learned confidence, designed to promote robustness against outliers. 
DELS-MVS is iterative, hence it does not require the construction of a large cost volume. 
Furthermore, no explicit discretization of the depth space within a min./max. range is required, as DELS-MVS explores the
epipolar line dynamically.
We confirmed the robustness of our approach through evaluations on the ETH3D~\cite{eth3d},  DTU \cite{dtu} and Tanks and Temples~\cite{tanksandtemples} benchmarks, where DELS-MVS achieves competitive results compared to the state-of-the-art. \textbf{Acknowledgement:} This work has been supported by the FFG, Contract No. 881844: "Pro$^2$Future".
\FloatBarrier

{\small
\bibliographystyle{ieee_fullname}
\bibliography{refs}
}

\end{document}